\documentclass[10pt,twocolumn,letterpaper]{article}

\usepackage{cvpr}
\usepackage{times}
\usepackage{epsfig}
\usepackage{graphicx}
\usepackage{amsmath}
\usepackage{amssymb}

\usepackage{pifont}
\usepackage{dsfont}
\usepackage{paralist}
\usepackage{multirow}
\usepackage{subcaption}
\usepackage[ruled]{algorithm2e}
\SetKwComment{Comment}{$\triangleright$\ }{}
\usepackage{booktabs}

\DeclareMathOperator*{\argmin}{arg\,min}

\usepackage{xcolor}

\usepackage[pagebackref=true,breaklinks=true,letterpaper=true,colorlinks,bookmarks=false]{hyperref}

\cvprfinalcopy % *** Uncomment this line for the final submission

\def\cvprPaperID{45} % *** Enter the CVPR Paper ID here

\ifcvprfinal\fi
\begin{document}

%%%%%%%%% TITLE
\title{Data from Model: Extracting Data from Non-robust and Robust Models}

\author{Philipp Benz\thanks{indicates equal contribution. Correspondence to {\tt\footnotesize pbenz@kaist.ac.kr} and {\tt\footnotesize chaoningzhang1990@gmail.com}  }
% Robotics and Computer Vision (RCV) Laboratory\\
% Korea Advanced Institute of Science and Technology (KAIST)\\
% 291 Daehak-ro, Yuseong-gu, Daejeon 34141, Korea\\
% {\tt\small pbenz@kaist.ac.kr}
% For a paper whose authors are all at the same institution,
% omit the following lines up until the closing ``}''.
% Additional authors and addresses can be added with ``\and'',
% just like the second author.
% To save space, use either the email address or home page, not both
\and
Chaoning Zhang$^{*}$
% Korea Advanced Institute of Science and Technology (KAIST)\\
% 291 Daehak-ro, Yuseong-gu, Daejeon 34141, Korea\\
% {\tt\small chaoningzhang1990@gmail.com}
\and 
Tooba Imtiaz
% {\tt\small timtiaz@kaist.ac.kr}
\and 
In-So Kweon 
\and 
% {\tt\small iskweon@kaist.ac.kr}\\
% {\small $^*$ indicates equal contribution }\\
% Robotics and Computer Vision (RCV) Laboratory\\
\\Korea Advanced Institute of Science and Technology (KAIST)
% 291 Daehak-ro, Yuseong-gu, Daejeon 34141, Korea\\
}

\maketitle
%\thispagestyle{empty}

%%%%%%%%% ABSTRACT
\begin{abstract}
The essence of deep learning is to exploit data to train a deep neural network (DNN) model. This work explores the reverse process of generating data from a model, attempting to reveal the relationship between the data and the model. We repeat the process of Data to Model (DtM) and Data from Model (DfM) in sequence and explore the loss of feature mapping information by measuring the accuracy drop on the original validation dataset. We perform this experiment for both a non-robust and robust origin model. Our results show that the accuracy drop is limited even after multiple sequences of DtM and DfM, especially for robust models. The success of this cycling transformation can be attributed to the shared feature mapping existing in data and model.
Using the same data, we observe that different DtM processes result in models having different features, especially for different network architecture families, even though they achieve comparable performance.
% learned and extracted during the DtM and DfM process. 
% With the feature mapping as the link between data and model, the two can be transformed into each other as we show through the DfM process. 
% Based on this feature mapping understanding we find that different DNN families can learn different feature mappings from the same training dataset. 
% This further provides insight into why the targeted transferability of adversarial examples fails. 
\end{abstract}

%%%%%%%%%%%% NOTES %%%%%%%%%%%%%%%%
% \CN{We need to show that different models learn different features because there are a lot of features. For models trained from the extracted dataset, containing only partial features, they can transfer better to each other.}
% Show results for 

%%%%%%%%% BODY TEXT
\section{Introduction}
\begin{figure}[t]
\centering
    \includegraphics[width=\linewidth]{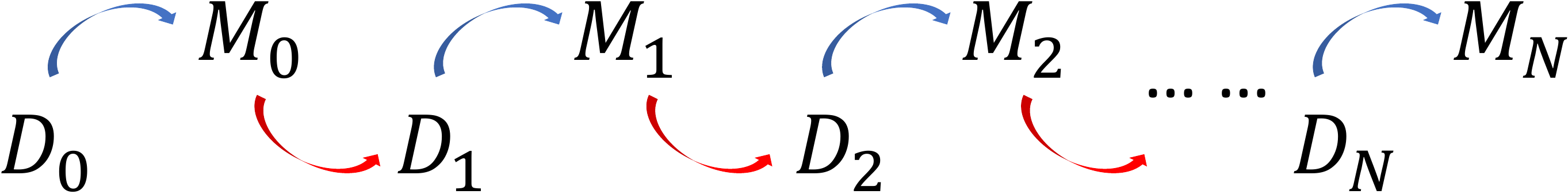}
\caption{The chain of performing DtM and DfM repetitively. The blue arrows indicate the DtM process and the red arrows indicate the DfM process.}
\label{fig:teaser}
\end{figure}

In deep learning applications, such as image classification~\cite{he2016identity,zhang2019revisiting}, data is used to train deep neural network (DNN) models. This work explores the reverse process of generating data from the model, with one general question in mind: What is the relationship between data and model? This question cannot be addressed well by only focusing on the model training process, Data To Model (DtM)~\cite{krizhevsky2012imagenet}. Thus we combine the widely adopted DtM with its reverse process of Data from Model (DfM). More specifically, we repeat the process of DtM and DfM in sequence and measure the accuracy over the original validation dataset. In this chaining process, we always assume access to only either the data or the model generated in the previous process. 
More specifically, in the DtM process, we can only access the data generated from the previous DfM process, and similarly, in the DfM process, we only access the model generated by the previous DtM process.
This chain process is depicted in Figure~\ref{fig:teaser}.

% Our intuition comes from one basic conjecture that a link exists between the data and the model, which is formulated as a feature mapping.
Our work is mainly inspired by~\cite{ilyas2019adversarial} which attributes the success of adversarial examples to the existence of non-robust features (mappings) in a dataset. During the typical DtM process, these non-robust features are learned by a model which consequently has the same non-robust features. This indicates the feature mapping as the link between data and model. 
% In image classification tasks, the dataset can be viewed as a container for a multitude of possible feature mappings from images to ground truth labels. 
% In DtM, the DNN exploits a subset of useful feature mappings to achieve the goal of classifying most images into the right category. 
% The DfM process is then able to extract an approximation of the features learned during the DtM process. 
% This dataset will not be the same as the original one since the model exploits only a subset of features instead of memorizing the exact data content. 
In their work, the original training dataset is adopted as the background image in the data extraction process~\cite{ilyas2019adversarial}. In this work, we explore the possibility of retrieving a learned feature mapping from a trained model without the original training dataset, which makes the DfM process more meaningful. Moreover, we iterate the DtM and DfM process in sequence instead of just performing it once. Another aspect of this work is to explore whether such feature mappings are the same or different for different runs for the same or different architectures. 

To decode the learned features of a model into a dataset without knowledge of the original training data, we adopt random substitute datasets as background to increase sample diversity and introduce virtual logits to model the logit behavior of DNNs. 
% Based on these virtual logits, feature images can be generated, which supersede the need of the original training dataset for training additional models. 
%The application of virtual logits distinguishes our work from the line of knowledge distillation (KD), since in KD commonly the output logits of a teacher model are used to train a student model.
Our experiments show that we can obtain models with similar properties as the original model in terms of both accuracy and robustness. We showcase the effectiveness of our approach on MNIST and CIFAR10 for both non-robust and robust origin models. %Given that the extracted dataset represents the learned features of a model, we further reveal that different models learn different sets of features. 

\section{Related Work}
DNNs are vulnerable to small, imperceptible perturbations~\cite{szegedy2013intriguing}. This intriguing phenomenon led to various attack~\cite{goodfellow2014explaining,carlini2017towards,madry2017towards,moosavi2017universal,zhang2019cd-uap} and defense methods~\cite{chakraborty2018adversarial}.
% , whereby attack methods still dominate this competition, with adversarial training~\cite{goodfellow2014explaining} being one of the most effective defense method. 
% Attack methods can be divided into two broad categories: image-dependant and universal attacks~\cite{moosavi2017universal,poursaeed2018generative,zhang2019cd-uap}. 
Interpretations for the reason of the existence of adversarial examples have been explored in~\cite{goodfellow2014explaining,tanay2016boundary,zhang2020understanding}. Ilyas~\etal~\cite{ilyas2019adversarial} attributed the phenomenon of adversarial examples to the existence of non-robust features. They introduce features as a mapping relationship from the input to the output. We adopt this definition and aim to extract this feature mapping from the model to the data. Model visualization methods~\cite{mordvintsev2015inceptionism,mahendran2015understanding} can be seen as a DfM method, without training a new model on the extracted data. Such methods are commonly further exploited for model compression~\cite{haroush2019knowledge,bhardwaj2019dream} techniques. Instead of compressing models,~\cite{wang2018dataset} aims to compress an entire dataset into only a few synthetic images. Training on these few synthetic images, however, leads to a serious performance drop.

\section{Methodology}
Given a $K$-classification dataset $\mathcal{D}$ consisting of data samples $x \in \mathds{R}^d$ and their corresponding true class $y \in [1, K]$, a DNN $\mathcal{M}_\theta$ ($\theta$ omitted from now on) parameterized through the weights $\theta$ is commonly trained via mini-batch stochastic gradient descent (SGD) to achieve $\argmin_\theta \mathds{E}_{(x,y)\sim \mathcal{D}} [\mathcal{L}(\theta, x, y)]$. 
%Here $\mathcal{L}(\theta, x, y)$ indicates the loss function, for which cross-entropy is a common choice. 
In this work, we term this process DtM (data to model) and we explore its reverse process of extracting data $\mathcal{D}'$ from model $\mathcal{M}$ (DfM).
% In this data to model (DtM) transformation, a set of feature mappings $\mathcal{F}$ given in the dataset is encoded in $\theta$. 
% Note that this set is likely only a subset of all available feature mappings in the dataset, and therefore multiple solutions can exist to learn feature mappings from data. 
% Ilyas \etal~\cite{ilyas2019adversarial} argued that in the case of standard model training the resulting feature mappings contain non-robust features, which leads to vulnerability of the DNN model to adversarial examples~\cite{szegedy2013intriguing}. 
% The goal of this work is to reverse the DtM process and retrieve the learned features from a model, or in other words to extract data $\mathcal{D}'$ from model $\mathcal{M}$ (DfM). 
% The DtM and DfM processes can be seen as counterparts and can, therefore, be applied consecutively. 
More specifically, starting from origin dataset $\mathcal{D}_0$ DtM results in origin model $\mathcal{M}_0$, with DfM, $\mathcal{D}_1$ can be extracted which leads to $\mathcal{M}_1$ through DtM and so on. 
% Through this chaining process, $\mathcal{M}_1$ and following models in the chain will exhibit similar properties as the origin model $\mathcal{M}_0$. 
During the DtM and DfM process, we assume having no access to the previous models and datasets, respectively. 

% Following the intuition of~\cite{ilyas2019adversarial} that adversarial examples are features, we conjecture that the features encoded as weights of a DNN can be decoded into a dataset through the adversarial crafting process. In particular, we deploy projected gradient descent (PGD)~\cite{madry2017towards} for feature extraction. 
To retrieve features from a model and store them in the form of data, we deploy the $l_2$-variant of projected gradient descent (PGD)~\cite{madry2017towards}.
Due to the absence of the original dataset, we leverage substitute images $x_s$ from a substitute dataset $\mathcal{D}_s$, as well as virtual logits $Z_v$ as the target values for the gradient-based optimization process. We specify the logit output of a classifier as $Z_\theta(\cdot)$ and use the $l_2$-loss between the network output logit and the virtual target logit $||Z_\theta(x_s) - Z_v||_2$ as the loss function optimized by PGD.
The retrieved dataset consists of images $x_s'=x_s+\delta$, where $\delta$ indicates a vector optimized through PGD and $x_s'$ lies in the range $[0,1]$. The data samples $x_s'$ and their respective output logit values $Z_\theta(x_s')$ represent the new dataset. 

After the DfM process, the retrieved dataset can be used in the DtM process, by training the model weights $\theta'$ with the $l_2$-distance between the previously stored ground truth logit vector and the output logit vector, $||Z_{\theta}(x_s') - Z_{\theta'}(x_s')||_2$.

We heuristically found a simple scheme of Gaussian distributed values $\mathcal{N}(\mu, \sigma)$ for the virtual logits. The highest logit is determined by $\mathcal{N}(20, 2)$, and the mean values of the remaining logit values are equally separated between $[-3,3]$ with $\sigma=1$. The order of logit values is chosen randomly to introduce diversity into the dataset.

In the above process, the origin model $\mathcal{M}_0$ is non-robust. Following~\cite{ilyas2019adversarial} we also use a robust model for $\mathcal{M}_0$ obtained with adversarial training and repeat the chaining process. 

% To obtain more robust feature mappings, adversarial training~\cite{goodfellow2014explaining} can be deployed following $\argmin_\theta \mathds{E}_{(x,y)\sim \mathcal{D}} [\max_{v \in \mathcal{S}} \mathcal{L}(\theta, x+v, y)]$,
% where $\mathcal{S}$ indicates a set of permissible perturbations $v$. The perturbation $v$ is restricted to obey the constraint $||v||_p \leq \epsilon$, where $||\cdot||$ indicates the $l_p$-norm and $\epsilon$ denotes the allowed perturbation magnitude.

\section{Experiments}

\subsection{DfM and DtM in sequence}
\begin{figure}[t]
\centering
    \includegraphics[width=0.49\linewidth]{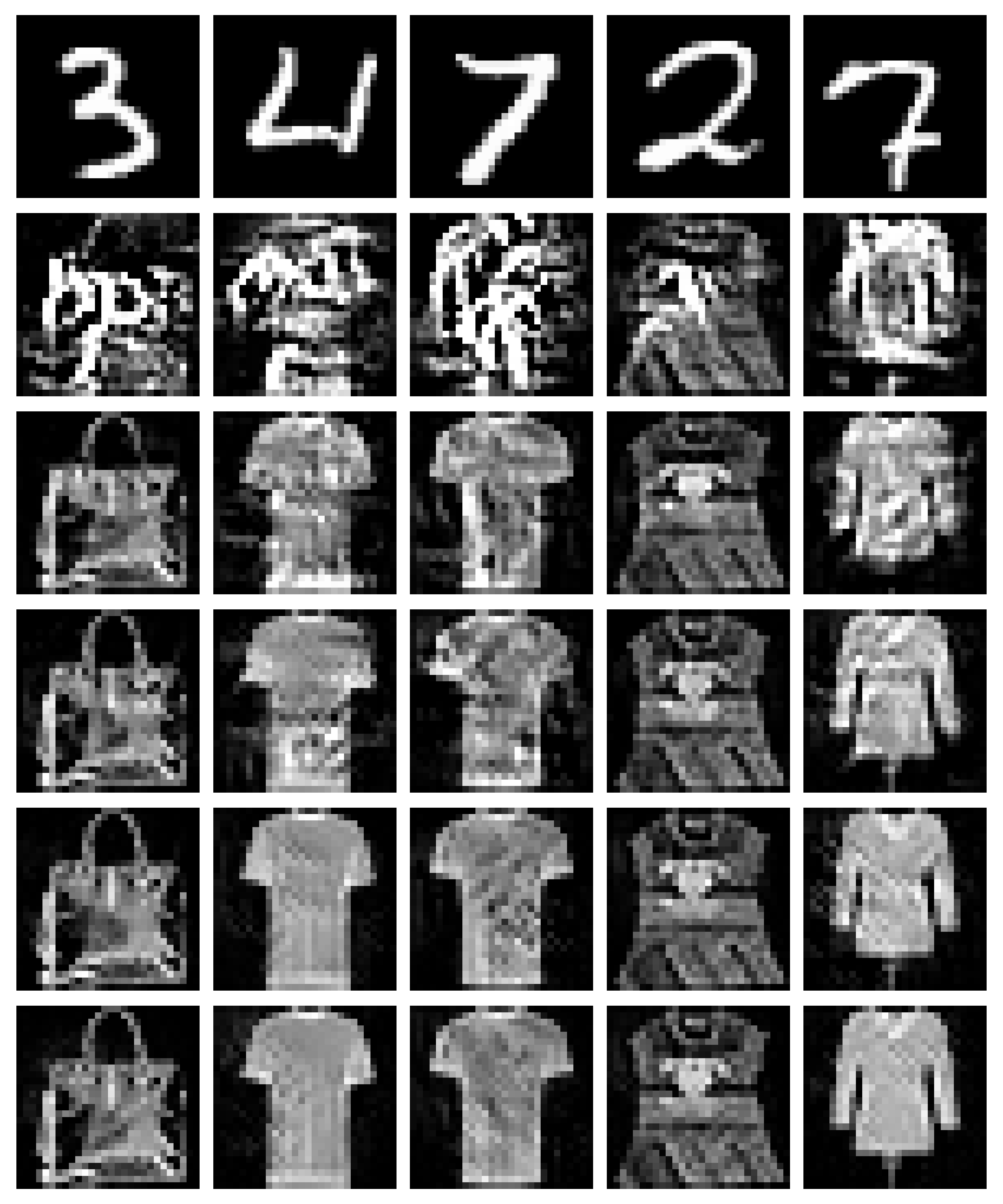}
    \includegraphics[width=0.49\linewidth]{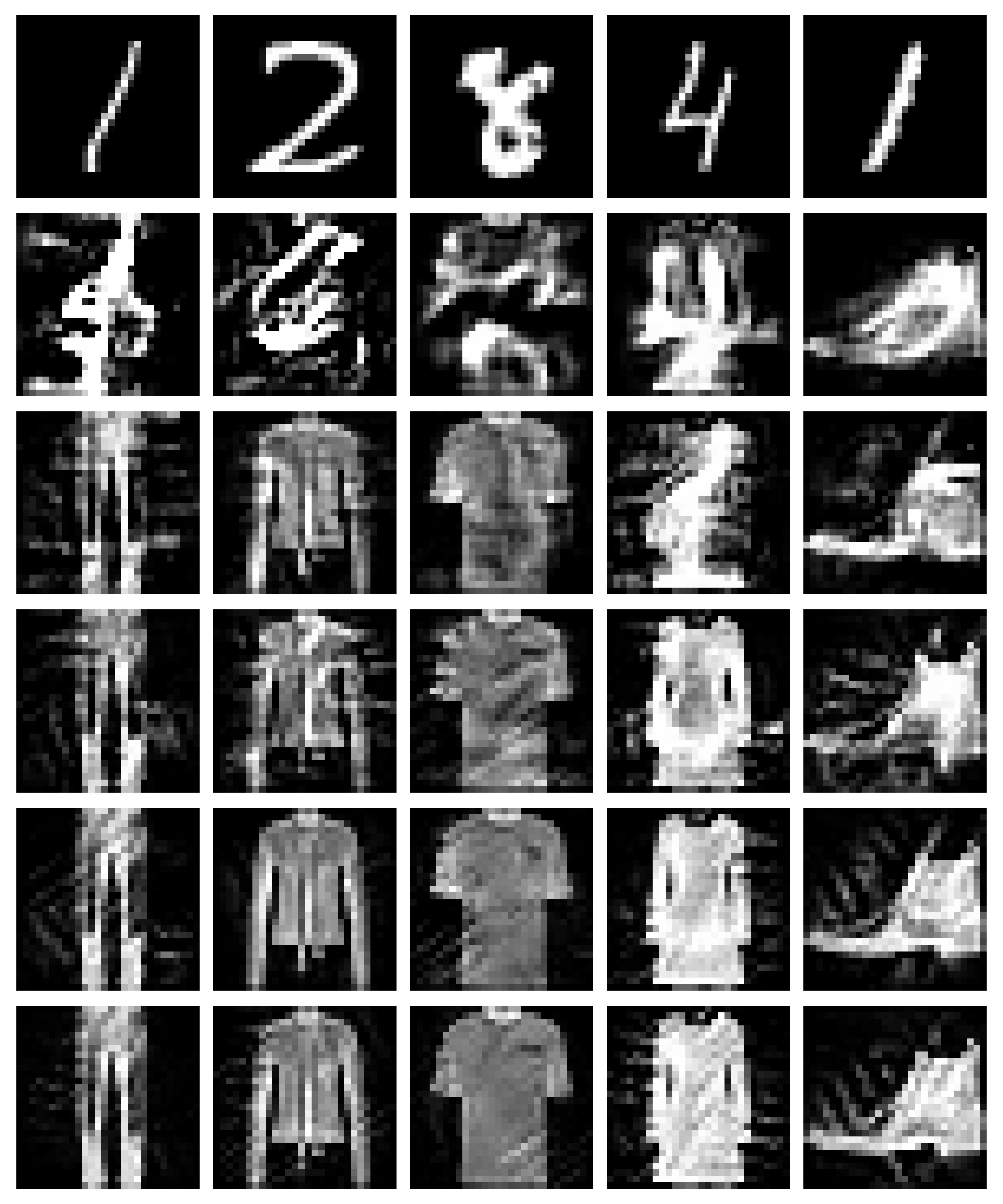}
\caption{Qualitative results for the DfM process starting from a non-robust origin model ($5$ columns on the left) and a robust origin model ($5$ columns on the right). The first row indicates the origin dataset. The subsequent rows indicate the obtained dataset after the $n$-th DtM and DfM process. The results are shown for the LeNet architecture with Fashion MNIST as the background images.}
\label{fig:qual_non_robust}
\end{figure}
We sequentially apply DfM and DtM starting from the origin model $\mathcal{M}_0$. The non-robust and robust origin models $\mathcal{M}_0$ are obtained through standard and adversarial training, respectively, on LeNet for MNIST and VGG8 for CIFAR10. VGG8 refers to a VGG network with only one convolution layer between each max pooling operation. To obtain $\mathcal{D}_i$ from the preceding model $\mathcal{M}_{i-1}$, $500$k images are extracted from $\mathcal{M}_{i-1}$ through the DfM process. For the background images, we choose Fashion MNIST and MS-COCO as background images for MNIST and CIFAR10, respectively. The generated images are shown in Figure~\ref{fig:qual_non_robust}. %We selected these two substitute datasets, since they do not share semantic features with the original datasets for the most part. 
In subsequent DtM, dataset $\mathcal{D}_i$ is then used to train the model $\mathcal{M}_i$ which is an independent model of the same architecture. The accuracy for all models is reported on the original validation dataset. We present the results with five repetitions of this process in Table~\ref{tab:chaining}.

\begin{table}[t]
\caption{Applying DtM and DfM in sequence for standard and adversarially trained models.}
\label{tab:chaining}
\centering
\small
\setlength\tabcolsep{2.5pt}
% \scalebox{0.9}{
    \begin{tabular}{c cc | cc }
    \toprule
    & \multicolumn{2}{c}{LeNet (MNIST)} & \multicolumn{2}{c}{VGG8 (CIFAR10)} \\ % & \multicolumn{2}{c}{VGG11 (CIFAR10)} \\
    & non-robust $\mathcal{M}_0$ & robust $\mathcal{M}_0$ & non-robust $\mathcal{M}_0$ & robust $\mathcal{M}_0$ \\% & non-robust & robust \\
    \midrule
    $\mathcal{M}_0$ & $99.5$ & $98.7$ & $92.2$ & $87.0$ \\% & $92.5$ \\
    $\mathcal{M}_1$ & $98.5$ & $97.7$ & $89.4$ & $88.1$ \\% & $90.4$ \\
    $\mathcal{M}_2$ & $96.6$ & $96.1$ & $80.1$ & $82.5$ \\% & $85.2$ \\
    $\mathcal{M}_3$ & $91.5$ & $95.2$ & $66.8$ & $71.7$ \\% & $76.7$ \\
    $\mathcal{M}_4$ & $87.4$ & $94.2$ & $52.5$ & $58.7$ \\% & $63.5$ \\
    $\mathcal{M}_5$ & $76.5$ & $93.7$ & $27.5$ & $44.8$ \\% & $44.4$ \\
    \bottomrule
    \end{tabular}
\end{table}

\begin{table}[t]
\caption{Robustness evaluation of the models obtained during the chaining process for a non-robust (left) and robust (right) origin model. The results are reported for the LeNet architecture on MNIST.}
\label{tab:robustness}
\centering
\small
\setlength\tabcolsep{2.5pt}
% \scalebox{0.9}{
    \begin{tabular}{c | ccccc | ccccc}
    \toprule
     & \multicolumn{5}{c}{non-robust $\mathcal{M}_0$} & \multicolumn{5}{c}{robust  $\mathcal{M}_0$} \\
     \midrule
     $\epsilon$         & $0$ & $1$ & $2$ & $3$ & $4$ & $0$ & $1$ & $2$ & $3$ & $4$ \\
     \midrule
     $\mathcal{M}_0$ & $99.5$ & $74.2$ & $4.5$ & $0.1$ & $0.1$ & $98.7$ & $91.0$ & $58.9$ & $10.6$ & $0.7$ \\
     $\mathcal{M}_1$ & $98.5$ & $61.3$ & $1.4$ & $0.0$ & $0.0$ & $97.7$ & $81.1$ & $25.4$ & $1.1$  & $0.0$ \\
     $\mathcal{M}_2$ & $96.6$ & $31.2$ & $0.2$ & $0.0$ & $0.0$ & $96.1$ & $71.2$ & $12.4$ & $0.1$  & $0.0$ \\
     $\mathcal{M}_3$ & $91.5$ & $6.5$ & $0.0$  & $0.0$ & $0.0$ & $95.2$ & $54.2$ & $3.24$ & $0.2$  & $0.1$ \\
     $\mathcal{M}_4$ & $87.4$ & $1.1$ & $0.0$  & $0.0$ & $0.0$ & $94.2$ & $31.8$ & $1.0$  & $0.1$  & $0.3$ \\
     $\mathcal{M}_5$ & $76.5$ & $0.0$ & $0.0$  & $0.0$ & $0.0$ & $93.7$ & $13.3$ & $0.6$  & $0.2$  & $0.2$ \\
    %  \midrule
    %  $\mathcal{M}_0$ & $98.7$ & $91.0$ & $58.9$ & $10.6$ & $0.7$ \\
    %  $\mathcal{M}_1$ & $97.7$ & $81.1$ & $25.4$ & $1.1$  & $0.0$ \\
    %  $\mathcal{M}_2$ & $96.1$ & $71.2$ & $12.4$ & $0.1$  & $0.0$ \\
    %  $\mathcal{M}_3$ & $95.2$ & $54.2$ & $3.24$ & $0.2$  & $0.1$ \\
    %  $\mathcal{M}_4$ & $94.2$ & $31.8$ & $1.0$  & $0.1$  & $0.3$ \\
    %  $\mathcal{M}_5$ & $93.7$ & $13.3$ & $0.6$  & $0.2$  & $0.2$ \\
    \bottomrule
    \end{tabular}
\end{table}

Qualitative results in Figure~\ref{fig:qual_non_robust} show that the extracted images look totally different from the original images, due to which it might be tempting to expect the models trained on them will work poorly on the original validation dataset. Table~\ref{tab:chaining}, however, shows that comparable performance is achieved and this is due to similar feature mappings existing in the generated images despite the large visual discrepancy. Nonetheless, we observe that there is a general trend that the accuracies for both the non-robust and robust models decrease for each sequence of DfM/DtM. The accuracy increase by $1.1\%$ from $\mathcal{M}_0$ to $\mathcal{M}_1$ for the robust VGG8 model is the only exception to this. 
For the non-robust origin model $\mathcal{M}_0$, the accuracy drop is trivial in the first few iterations of the DfM and DtM process and it becomes more observable in later iterations. For the robust origin model $\mathcal{M}_0$, the accuracy is retained better. For example, the robust LeNet only decreases by $5\%$ from $\mathcal{M}_0$ to $\mathcal{M}_5$, while the non-robust LeNet decreases by $23\%$. 
% Overall, these results indicate that the DfM and DtM process can successfully be repeated multiple times and that it is, therefore, possible to retrieve the features learned through a DNN and transform them again into a dataset.

We further investigate the robustness to adversarial examples for the models from the DfM/DtM process. Therefore we evaluate the different retrieved MNIST models on adversarial examples generated with $l_2$-PGD under different $\epsilon \in \{0,4\}$, $20$ update steps and the corresponding step size calculated as $2.5\epsilon/\text{steps}$. 
Similar to the observed accuracy drop for clean images, a similar trend occurs when the model is under attack. However, the accuracy degradation seems to be more severe for robustness. For example, the accuracy of the non-robust LeNet drops from $74.2\%$ to $0.0\%$ for a relatively weak attack of $\epsilon=1$ while the clean image accuracy drops from $99.5\%$ to $76.5\%$. Similar behavior is observed for the models originating from the robust $\mathcal{M}_0$.
%We conjecture that this more serious drop is caused the transformation of robust features into non-robust features.
It is worth mentioning that the models originating from the robust $\mathcal{M}_0$ are consistently more robust than their counterparts originating from the non-robust $\mathcal{M}_0$. After two subsequent DfM iterations starting from the robust $\mathcal{M}_0$, $\mathcal{M}_2$ still shows similar robustness as the standard non-robust model $\mathcal{M}_0$. 
This result suggests that DfM can be applied as an alternative to adversarial training.
% From the results in Table~\ref{tab:robustness} we observe that the models in the chain originating from the non-robust $\mathcal{M}_0$ decrease rapidly in their robustness to adversarial examples. $\mathcal{M}_3$ only achieves an accuracy of $6.5\%$ for a relatively weak attack of $\epsilon=1$. The models originating from the robust $\mathcal{M}_0$ exhibit similar performance to $\mathcal{M}_0$ and continue to show higher robustness to adversarial examples, compared to their counterparts originating from the non-robust $\mathcal{M}_0$. After two subsequent DfM iterations starting from the robust $\mathcal{M}_0$, $\mathcal{M}_2$ still shows similar robustness as the standard non-robust model $\mathcal{M}_0$. 
% As expected, the accuracy of the standard trained model decreases drastically for increasing epsilon. For the models obtained through the chaining process originating from the adversarially trained model, higher robustness can be observed. This result suggests that DfM can be applied as an alternative to adversarial training. 

% Only generating D1.
% Higher complexity models also in better performance 
% D1(VGG16) --> M(D1) --> M(CIFAR10)
\subsection{DfM and DtM on different architectures}

\begin{table}[t]
\caption{Cross-training of the extracted datasets from non-robust (top) and robust (bottom) models. The models were originally trained on CIFAR10. The robust models were adversarially trained with the $l_2$ variant of PGD. The rows indicate the model from which the data was extracted. The columns indicate the trained model. The values indicate the accuracy of the CIFAR-10 test dataset.}
\label{tab:non-robust-cross-train}
\centering
\small
\setlength\tabcolsep{1.5pt}
% \scalebox{0.9}{
    \begin{tabular}{ccccccc}
    \toprule
                     & & VGG16 & VGG19 & ResNet18 & ResNet50 \\
    \midrule
    \parbox[t]{3mm}{\multirow{3}{*}{\rotatebox[origin=c]{90}{non-rob.}}} 
    & VGG16 ($93.8$)    & $89.6$ & $90.1$ & $89.9$ & $90.3$ \\
    & VGG19 ($93.6$)    & $89.7$ & $90.1$ & $90.6$ & $90.3$ \\
    & ResNet18 ($95.1$) & $87.9$ & $88.0$ & $89.7$ & $89.6$ \\
    \midrule
    \parbox[t]{3mm}{\multirow{3}{*}{\rotatebox[origin=c]{90}{robust}}} 
    & VGG16 ($88.7$)    & $90.3$ & $90.5$ & $90.3$ & $90.5$ \\
    & VGG19 ($87.6$)    & $87.9$ & $88.0$ & $88.0$ & $88.1$ \\
    & ResNet18 ($90.2$) & $91.3$ & $91.1$ & $91.6$ & $91.5$ \\
    \bottomrule
    \end{tabular}
\end{table}

The above analysis shows that DtM and DfM can be performed for the same and simple architecture with a limited performance drop. 
% Here we show that this process is not limited to this setup and that the performance drop is also not necessarily observed. 
Here we apply DfM to the standard and adversarially trained CIFAR10 models and train different state-of-the-art architectures on the extracted data. For simplicity we stop the chaining process after obtaining $\mathcal{M}_1$. The results in Table~\ref{tab:non-robust-cross-train} show that all model architectures can be successfully trained on the extracted data. Similar to Table~\ref{tab:chaining}, a performance drop is observed for the data extracted from the non-robust $\mathcal{M}_0$.
%For example the ResNet50 architecture trained on the data extracted from non-robust VGG16 decreases by $3.5\%$.
For the robust $\mathcal{M}_0$, however, we observe that the retrained $\mathcal{M}_1$ consistently outperforms their corresponding $\mathcal{M}_0$ for both similar and different architectures, which is somewhat surprising. 
% Similar to the performance improvement from the robust $\mathcal{M}_0$ to $\mathcal{M}_1$ on CIFAR10 in Table~\ref{tab:chaining}, we observe that the models trained on the extracted data from the robust origin models even outperform the origin models. 
% We conjecture that this is due to the diversity introduced through the vast amount of data, which DfM allows to extract.
% The shown results suggest that one model architecture can be successfully trained on the extracted data from another model.

\subsection{Do different models learn different feature mappings?}
\begin{table}[t]
\caption{Cross-evaluation of datasets extracted from non-robust (top) and robust (bottom) models. The models were originally trained on CIFAR-10. The robust models were obtained with adversarial training via the $l_2$ variant of PGD. The diagonal values were obtained for the same architecture but a different training run. The accuracy of the extracted data on the original model is $100\%$.}
\label{tab:non-robust-cross-eval}
\centering
\small
\setlength\tabcolsep{1.5pt}
% \scalebox{0.9}{
    \begin{tabular}{ccccccc}
    \toprule
            & & VGG16 & VGG19 & ResNet18 & ResNet50 \\
    \midrule
    \parbox[t]{3mm}{\multirow{4}{*}{\rotatebox[origin=c]{90}{non-robust}}} 
    & VGG16    & $43.2$ & $40.4$ & $36.5$ & $30.5$ \\
    & VGG19    & $50.1$ & $48.7$ & $45.4$ & $37.8$ \\
    & ResNet18 & $36.9$ & $34.9$ & $55.2$ & $43.8$ \\
    & ResNet50 & $41.3$ & $40.6$ & $60.0$ & $62.6$ \\
    \midrule
    \parbox[t]{3mm}{\multirow{4}{*}{\rotatebox[origin=c]{90}{robust}}} 
    & VGG16    & $48.5$ & $43.6$ & $45.4$ & $44.0$ \\
    & VGG19    & $38.3$ & $38.9$ & $36.9$ & $36.0$ \\
    & ResNet18 & $42.0$ & $37.8$ & $50.1$ & $47.7$ \\
    & ResNet50 & $35.4$ & $31.9$ & $39.7$ & $35.7$ \\
    \bottomrule
    \end{tabular}
\end{table}

Given the possibility to extract a certain feature mapping from a model, in this section, we analyze whether different models trained from the same origin dataset $\mathcal{D}_0$ have different feature mappings. To this end, we utilize $10$k feature images extracted from models trained under the same conditions, and perform a cross-evaluation of the model accuracy on each other. The results for models with standard and adversarial training are reported in Table~\ref{tab:non-robust-cross-eval}. We observe that the cross-evaluation accuracies are higher than random guess, which indicates that some shared feature mappings are learned. However, for both non-robust and robust models only in a few cases an accuracy higher than $50\%$ is achieved. This phenomenon can also be observed when the extracted dataset was evaluated on an independently trained instance of the same architecture as the original architecture. 
% The low performances on the cross-evaluation shows that the extracted dataset from one classifier yields relatively low accuracy on another classifier, even on an independent instance of the same model architecture trained from scratch. 
The relatively low cross-evaluation accuracies illustrate that models from different runs learn different features which are not fully compatible with each other.

% This observation indicates why targeted adversarial examples do not transfer across models in terms of targeted accuracy. The reason is that the crafted examples mainly attack features learned by the initially targeted model, which are simply not existent in another model and can therefore not be compromised.
% Table~\ref{tab:targeted_transferabilty_ae} shows the targeted success rate of adversarial examples generated for one model architecture and cross evaluated on other architectures. The targeted accuracy refers to the accuracy that the adversarial example is classified as the intended target class. In the case of image-dependant adversarial examples, the accuracy for the initially generated models is $100\%$, while the targeted success rate for the other architectures is nearly zero. 
Another interesting observation is that model architectures from the same network family, VGG family for instance, seem to have more common feature mappings than different architectures. For example, for the feature images extracted from the standard ResNet50, the ResNet networks exhibit an accuracy of around $60\%$, while the VGG networks only show an accuracy of around $40\%$. This phenomenon is more prevalent in non-robust models than in robust models. Overall, the results show that different models learn different feature mappings from the same dataset even though they have comparable classification accuracy.

\section{Conclusion}
In this work we introduced the Data from Model (DfM) process, a technique to reverse the conventional model training process, by extracting data back from the model. A model trained on the generated dataset that look totally different from the original dataset can achieve comparable performance as their counterparts trained on the origin dataset. The success of this technique confirmed feature mapping as the link between data and model. Our work provides insight about the relationship between data and model as well understanding of model robustness. %This insight provided us with a framework to analyze the learned features of a DNN. We provided evidence that different model architectures learn different non-robust and robust features, but do also have the capability to learn the same features. 
% This insight lead us to the explanation of why targeted adversarial examples do not transfer. 
% However, we provided evidence that targeted universal perturbations show transferability among their network families.

{\small
\bibliographystyle{ieee_fullname}
\bibliography{data_from_model}
}

\end{document}